\documentclass[letterpaper, 10 pt, conference]{style}
\IEEEoverridecommandlockouts
\usepackage{cite}
\usepackage{amsmath,amssymb,amsfonts}
\usepackage[colorlinks=true, linkcolor=blue, citecolor=blue, urlcolor=blue]{hyperref}
\usepackage{algorithmic}
\usepackage{graphicx}
\usepackage{textcomp}
\usepackage{booktabs}
\usepackage{amsmath} 
\usepackage{tabularx} 
\usepackage{xcolor}
\def\BibTeX{{\rm B\kern-.05em{\sc i\kern-.025em b}\kern-.08em
    T\kern-.1667em\lower.7ex\hbox{E}\kern-.125emX}}

\begin{document}

\title{\LARGE \bf Network-Wide Traffic Volume Estimation from Speed Profiles using a Spatio-Temporal Graph Neural Network with Directed Spatial Attention \\

}

\author{Léo Hein$^{1,2}$, Giovanni De Nunzio$^{1}$, Giovanni Chierchia$^{2}$, Aurélie Pirayre$^{1}$, Laurent Najman$^{2}$
\thanks{$^{1}$IFP Energies nouvelles, 1 et 4 avenue de Bois-Préau, 92852 Rueil-Malmaison, France,
        {\footnotesize \{leo.hein, aurelie.chataignon, giovanni.de-nunzio\}@ifpen.fr}}%
\thanks{$^{2}$Univ Gustave Eiffel, CNRS, LIGM, Marne-la-Vallée, 77454, France, {\footnotesize \{laurent.najman, giovanni.chierchia\}@esiee.fr}}}%

\maketitle

\begin{abstract}
Existing traffic volume estimation methods typically address either forecasting traffic on sensor-equipped roads or spatially imputing missing volumes using nearby sensors. While forecasting models generally disregard unmonitored roads by design, spatial imputation methods explicitly address network-wide estimation; yet this approach relies on volume data at inference time, limiting its applicability in sensor-scarce cities. Unlike traffic volume data, probe vehicle speeds and static road attributes are more broadly accessible and support full coverage of road segments in most urban networks. In this work, we present the Hybrid Directed-Attention Spatio-Temporal Graph Neural Network (HDA-STGNN), an inductive deep learning framework designed to tackle the network-wide volume estimation problem. Our approach leverages speed profiles, static road attributes, and road network topology to predict daily traffic volume profiles across all road segments in the network. To evaluate the effectiveness of our approach, we perform extensive ablation studies that demonstrate the model’s capacity to capture complex spatio-temporal dependencies and highlight the value of topological information for accurate network-wide traffic volume estimation without relying on volume data at inference time.
\end{abstract}


\section{Introduction}
Estimating accurate traffic volume across an urban road network is a critical task for urban planning, traffic management, and environmental monitoring, such as air quality assessment \cite{b1}. Traffic volume can be precisely measured using sensor data collected from instrumented road segments. However, the installation and maintenance of such infrastructure are costly, resulting in sparse coverage in most urban networks and motivating the development of models to estimate traffic volume on unmonitored roads.

Two main directions have emerged in the traffic volume prediction literature: traffic volume forecasting, which predicts future volumes on sensor-equipped roads using past time series data; and spatial volume imputation, which estimates current volumes on unmonitored roads based on available sensor data in the network. While forecasting methods generally ignore non-instrumented roads, spatial imputation directly addresses the challenge of network-wide volume estimation. However, both approaches typically require volume data, at least partially, during inference, which limits their applicability in data-scarce environments and renders them ineffective on fully unmonitored networks.

To overcome this limitation, some studies have explored estimating traffic volume using more accessible descriptors. In particular, geospatial data providers offer widespread access to probe speed data and static road segment attributes. Many approaches build on traffic flow theory, which describes macroscopic relationships between key traffic state variables such as speed, density, and volume, usually leveraging fundamental diagrams (FDs) to translate speed into volume on specific road segments \cite{b2}. While being an active domain of research, FD-based approaches often struggle under free-flow conditions due to the limited descriptive power of constant speed profiles, as noted in \cite{b3}. 

In order to reduce modeling biases, some machine learning approaches have attempted to estimate volume from probe vehicle speed data, showing encouraging results \cite{b4, b5}. Yet, these models often ignore network topology and spatial dependencies, operating locally on individual road segments. This limits their predictive performance, since other studies have demonstrated the strong influence of topological context on the speed-flow macroscopic relation \cite{b6}.

Road networks can naturally be represented as fixed-structure temporal graphs, where node or edge attributes evolve over time. Spatio-Temporal Graph Neural Networks (STGNNs) have thus achieved state-of-the-art performance in capturing complex spatio-temporal dependencies in traffic states, particularly for forecasting tasks \cite{b7, b8, b9}, with many advanced architectures proposed. However, most models operate on sensor graphs created with artificial metrics between sensor stations, rather than on the road network’s topological graph. This choice aggregates local traffic behavior at a coarser scale and limits the understanding of flow dynamics, leading to models that may generalize poorly to unseen roads or networks.

Furthermore, most graph-based traffic prediction models are inherently highly network-specific and struggle to generalize to unseen topologies \cite{b10}, due to a lack of inductive capacity \cite{b11, b12}. An inductive model would learn a transformation based on node-level features and local structural patterns, allowing it to be applied to entirely new graphs or nodes without retraining or architectural modifications. In contrast, many existing spatio-temporal models exhibit transductive behavior: they rely on components tied to the specific training graph structure. 

While existing studies have addressed network-wide traffic volume estimation through spatial imputation using probe vehicle speed data, often incorporating topological information via graph-based approaches \cite{b13, b14, b15, b16, b17}, they are limited in their generalization capabilities and are inapplicable to sensor-free road networks. To fill this gap, we propose a novel deep-learning approach for network-wide traffic volume estimation, tackling the aforementioned challenges:

\begin{itemize}
    \item The dependence on volume data at inference is removed by learning to map daily speed profiles, static attributes, and local topology to daily traffic volumes.
    \item The model captures accurate local directional traffic dynamics by leveraging the topological graph of the road network and traffic flow orientation.
    \item The learned transformation relies only on local features, without any global parameters specific to the training network, making the model fully inductive and applicable to unseen road sections.
\end{itemize}

Specifically, we introduce the \textbf{HDA-STGNN} architecture, a hybrid two-branch learning framework that optimally processes both types of graph-based input data. The model integrates Directed Graph Attention (DGAT) layers to inductively capture directionally-oriented traffic flow patterns within local road network neighborhoods, while employing 1D temporal convolutions to model temporal dependencies. We evaluate our model on a real-world traffic volume estimation dataset from Lyon (2021), which combines traffic count data, GPS traces, and geospatial information to reconstruct the city’s topological road network graph. Extensive ablation studies highlight the contribution of each architectural component, demonstrating promising performance in estimating traffic volumes on previously unseen roads across the network.

\begin{figure*}[t] 
    \centering
    \includegraphics[trim=0cm 11.5cm 0cm 0cm, clip, width=\textwidth]{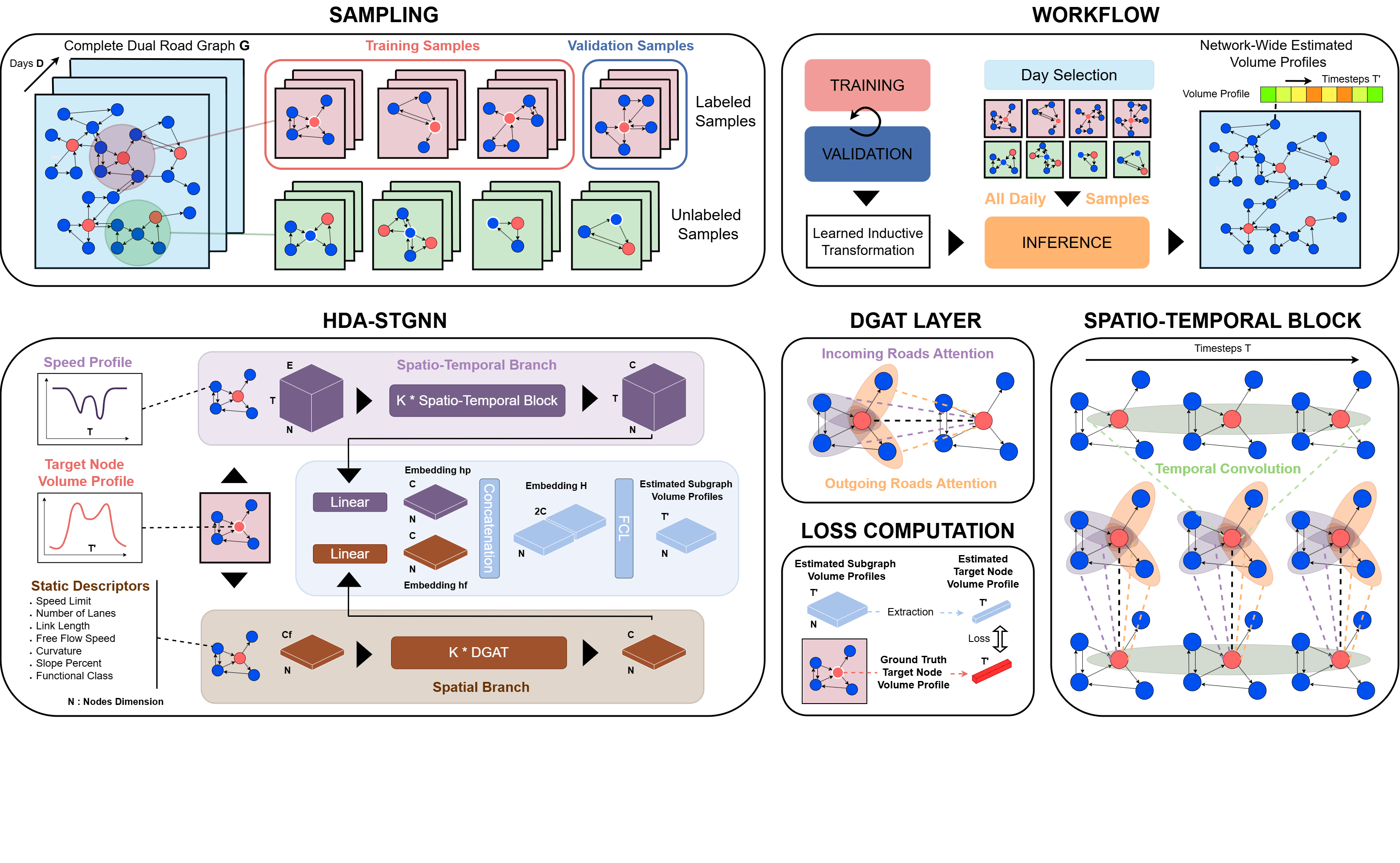}
    \caption{HDA-STGNN framework and architecture.}
    \label{Fig:main}
\end{figure*}

\section{Problem Definition}

We define our task as a particular case of traffic state estimation, where the objective is to learn a transformation from speed profiles and static road descriptors within a local topological context, into a representative daily traffic volume profile, applicable to any given road segment in a road network and for any given day. This transformation is learned in a supervised setting using real volume measurements, which are available only on a subset of labeled road segments, while descriptors and speed data are available for all road sections in the network.

We represent the topological road network as an oriented dual graph \( G = (V, E) \), where each node \( v \in V \) corresponds to a road segment, and edges in \(E \) represent physical connections between segments, taking into account all traffic rules. Each distinct road orientation is modeled as a separate node, and each edge is directed according to the traffic flow direction in the original (primal) road network. In this setting, the dual graph is a directed, node-attributed graph, where edges are structural and featureless, encoding absolute directional connectivity between road segments. This formulation enables a more natural use of message-passing operations for updating road segment representations. Moreover, the dual graph representation also allows for a better integration of traffic regulations, making it easier to account for constraints like one-way streets.

We denote by \( V_l \subset V \) the set of labeled nodes with observed traffic volumes and by \( V_n \subset V \) the unlabeled ones. Let \( F \in \mathbb{R}^{|V| \times C_f} \) be the static descriptors matrix associated to $G$, describing road segment characteristics, where \( C_f \) is the number of descriptors, and \( P \in \mathbb{R}^{|V| \times T \times D} \) be the tensor of speed profiles associated to $G$, with \( T \) considered time steps per day and \( D \) available days. Finally, \( Q \in \mathbb{R}^{|V_l| \times T' \times D} \) contains the observed traffic volumes on the equipped road segments, i.e., for all labeled nodes in $V_l$, for the same $D$ days, with $T'$ time steps per day. 

By denoting the $K$-hop neighborhood of a node $v$ as $\mathcal{N}_K(v)$, (i.e., all nodes within distance $k\leq K$, including $v$) we construct, for any labeled node $v_l \in V_l$ and day \( d \in \{1, \ldots, D\} \), each training sample as:
\begin{equation}
\left( G_{\mathcal{N}_K(v_l)},\; F_{\mathcal{N}_K(v_l)},\; P_{\mathcal{N}_K(v_l), d},\; Q_{v_l, d} \right)
\tag{$\star$}
\label{Eq:sample}
\end{equation}
where:
\begin{itemize}
    \item \( G_{\mathcal{N}_K(v_l)} \) is the $K$-hop neighborhood subgraph of target node $v_l$,
    \item \( F_{\mathcal{N}_K(v_l)} \in \mathbb{R}^{|\mathcal{N}_K(v_l)| \times C_f} \) contains all static descriptors of the local subgraph,
    \item \( P_{\mathcal{N}_K(v_l), d} \in \mathbb{R}^{|\mathcal{N}_K(v_l)| \times T} \) contains all speed profiles of the local subgraph for day \( d \),
    \item \( Q_{v_l, d} \in \mathbb{R}^{T'} \) is the ground-truth volume profile for node \( v_l \) on day \( d \).
\end{itemize}

Our supervised learning objective is to learn a function \( f_\theta \), where $\theta$ denotes learnable parameters, to estimate the daily volume profile of any $v_l \in V_l$ and \(
d\in \{1, \ldots, D\} \), such that:
\[
\hat{Q}_{v_l, d} = f_\theta(G_{\mathcal{N}_K(v_l)},\; F_{\mathcal{N}_K(v_l)},\; P_{\mathcal{N}_K(v_l), d}).
\]

Ultimately, our goal is to apply the learned function to every road segment in the network for any given day, whether within or beyond the working time range, by learning a generalizable transformation that is both spatially and temporally inductive, free from global or graph-specific parameters. Since the descriptors used during training are available for all nodes in the network, the model can be naturally applied to unlabeled segments within the same network. Furthermore, it is potentially transferable to entirely different road networks without architectural modifications or retraining, relying solely on its spatial generalization capacity in an out-of-distribution setting.

\section{Methodology}

\subsection{Proposed Model Architecture}

In this section we elaborate on the proposed \textbf{Hybrid Directed-Attention Spatio-Temporal Graph Neural Network (HDA-STGNN)} architecture. As shown in Figure \ref{Fig:main}, the model features two branches tailored to the two input tensors for the considered neighborhood subgraph: \(F_{\mathcal{N}_K(v_l)}\) and \(P_{\mathcal{N}_K(v_l), d}\), respectively representing the associated static descriptors and temporal speed data for a given day $d$. The spatio-temporal branch employs factorized blocks combining a Directed Graph Attention (DGAT) layer between two symmetric 1D temporal convolution (1D-CNN) layers to capture complex dependencies in the spatio-temporal speed profiles graph, while the spatial branch applies DGAT layers on the static descriptor graph to capture topology-aware and time-independent spatial dependencies. The complementary embeddings obtained from both branches are then concatenated and fed to fully connected layers to predict the daily volume profiles in the subgraph. Finally, we select the estimated volume profile of the sample target node $v_l$ to compute the loss function and train the model. Details of each component are described in the following sections.

\subsection{Directed Graph Attention Layers for Spatial Aggregation}
Graph Attention Networks (GATs)~\cite{b18} are widely used for learning on graph-structured data, such as road networks, and often outperform Graph Convolutional Networks (GCNs)~\cite{b19} by leveraging attention mechanisms that assign varying importance to neighboring nodes. This allows for more flexible and context-sensitive aggregation of information.

In a standard GAT, each node \( v \) aggregates messages from its 1-hop neighbors \( u \in \mathcal{N}_1(v) \) using an attention mechanism defined as follows:
\begin{align}
    e_{vu} = \text{LeakyReLU} \left( a^\top \left[ W \mathbf{h}_v \, \| \, W \mathbf{h}_u \right] \right),
\end{align}
where \( \mathbf{h}_v \in \mathbb{R}^C \) and \( \mathbf{h}_u \in \mathbb{R}^C \) denote the input feature vectors of nodes \( v \) and \( u \), respectively, with $C$ being the hidden dimension of the model, \( W \in \mathbb{R}^{C \times C} \) is a learnable projection matrix, \( a \in \mathbb{R}^{2C} \) is a learnable attention vector, and \( \| \) denotes vector concatenation. The scalar \( e_{vu} \) represents the raw attention score, indicating the relative importance of node \( u \)'s features when updating node \( v \)'s representation.

These scores are then normalized across the 1-hop neighborhood of node \( v \) using a softmax function:
\begin{align}
    \alpha_{vu} = \frac{\exp(e_{vu})}{\sum_{i \in \mathcal{N}_1(v)} \exp(e_{vi})},
\end{align}
and the new node representation is computed as a weighted sum of its neighbors' features, followed by a non-linear activation function \( \sigma \):
\begin{align}
    \mathbf{h}_v^{\text{new}} = \sigma \left( \sum_{u \in \mathcal{N}_1(v)}  \alpha_{vu} \left( W \mathbf{h}_u \right) \right).
\end{align}

While GATs support asymmetric node aggregation through the concatenation operation, they do not explicitly model edge direction, which limits their effectiveness on directed graphs. In road networks, where upstream and downstream connections both carry important but distinct semantic meaning, standard GATs often fall short, either by focusing solely on incoming edges in directed graphs, missing crucial traffic state information, or by overlooking directional patterns, indistinguishable in an undirected setting. As a result, traditional GATs fail to capture the spatial causality inherent in traffic volume dynamics.

To capture such directional dependencies in the oriented dual road network graph, we adopt Directed Graph Attention (DGAT) layers, based on \cite{b20, b21}, which extends standard GAT layers by explicitly encoding edge orientation. Each edge is assigned a label $dir$,  relative to the updated node, \textit{incoming}, \textit{outgoing} or \textit{self-loop}, associated with separate learnable parameters: a projection matrix \( W^{\text{dir}} \), an attention vector \( a^{\text{dir}} \), and a learnable directional embedding \( \mathbf{d}^{\text{dir}} \). The directed attention score is computed as:
\begin{align}
    e_{vu}^{\text{dir}} = \text{LeakyReLU} \left( a^{\text{dir}^\top} [W^{\text{dir}} \mathbf{h}_v \, \| \, W^{\text{dir}} \mathbf{h}_u \, \| \, \mathbf{d}^{\text{dir}}] \right),
\end{align}
and the updated node representation is obtained \textit{via} direction-aware weighted aggregation:
\begin{align}
    \mathbf{h}_v^{\text{new}} = \sigma \left( \sum_{u \in \mathcal{N}_1(v)} \alpha_{vu}^{\text{dir}} \left( W^{\text{dir}}\mathbf{h}_u \right) \right), \label{eq:dgat0}
\end{align}
where \( \alpha_{vu}^{\text{dir}} \) is the softmax-normalized directed attention coefficient computed just as in the standard GAT layer:

\begin{align}
    \alpha_{vu}^{\text{dir}} = \frac{\exp(e_{vu}^{\text{dir}})}{\sum_{i \in \mathcal{N}_1(v)} \exp(e_{vi}^{\text{dir}})}.
\end{align}

Directional attention vectors \( a^{\text{dir}} \) and projection weights \( W^{\text{dir}} \) enable the model to treat both edge directions distinctly, as it aggregates neighbors' information relatively to their connection to the node being updated. The model is thus able to distinguish between traffic flow states on upstream and downstream road segments, even when these segments share similar features, by leveraging directional context. In addition, the directional embedding vector $d^{\text{dir}} \in \mathbb{R}^{C}$ is a learned embedding that encodes the direction of an edge with respect to the updated node, with three possible types: \textit{incoming}, \textit{outgoing} or \textit{self-loop}. Each direction is associated with a dense vector of fixed dimension $C$, which is learned during training. During the forward pass, this vector is retrieved for each edge based on its label \textit{dir}, and concatenated with the projected representations of the source and target nodes. This concatenation enhances standard GAT operations by including both node identities and the direction of their connection, creating a richer vector to compute the attention score and allowing the model to better modulate message importance according to the structural role of each interaction in the graph.

\subsection{Processing Speed Profiles with a Spatio-Temporal Branch}

The spatio-temporal branch takes as input the tensor $P_{\mathcal{N}_K(v_l), d} \in \mathbb{R}^{|\mathcal{N}_K(v_l)| \times T \times E}$,  which represents the speed profile data as a spatio-temporal tensor with three distinct dimensions: spatial, temporal, and feature-wise. The feature dimension $E$ includes raw speed values along with a temporal encoding of hours and weekdays. This format is essential for capturing complex spatio-temporal dependencies, as commonly adopted in recent STGNN models. To handle these dependencies, we design a factorized spatio-temporal block by stacking spatial and temporal aggregation layers, inspired by related studies \cite{b22, b23}. 

While Recurrent Neural Networks (RNNs) are commonly used to model temporal dependencies, their complex mechanisms can be computationally expensive and impose strict causality constraints. In contrast, non-causal models such as Transformer encoders and 1D Convolutional Neural Networks (1D-CNNs) have gained popularity due to their flexibility. We justify the use of a non-causal model, as both input and output temporal windows correspond to the same day, allowing information from both past and future time steps to be equally relevant. Among non-causal models, 1D-CNNs are particularly well-suited for capturing short-range temporal patterns with lower computational cost than Transformers, making them an efficient choice for our task and more aligned with the variability of our application data.

Let’s consider the input embedding of the spatio-temporal block for layer \( k > 1 \), denoted as \( h^{(k-1, 0)} \in \mathbb{R}^{|\mathcal{N}_K(v_l)| \times T \times C} \) with \( C \) being the hidden channel dimension and \( h^{(0, 0)} = P_{\mathcal{N}_K(v_l), d} \in \mathbb{R}^{|\mathcal{N}_K(v_l)| \times T \times E} \) the initial speed features.

The following three operations are performed sequentially within each layer's block \( k \in \{1, \ldots, K\} \) over both the temporal and spatial dimensions of the embeddings: for all nodes \( v \in \mathcal{N}_K(v_l) \) and for all time steps \( t \in \{1, \ldots, T\} \).

First, a symmetric 1D-CNN is applied along the temporal dimension as follows:
\begin{align} 
h_{v,t}^{(k-1, 1)} = \sigma \left( \sum_{i=0}^{p-1} W_i^{(k,0)} \, h^{(k-1, 0)}_{v, t + i - \left\lfloor \frac{p}{2}\right\rfloor}\right) \label{eq:stgnn1},
\end{align}
where \( h_{v,t}^{(k-1, 1)} \in \mathbb{R}^C \),  \( p \) is the kernel size and $W_i^{(k,0)}$ $\in \mathbb{R}^{C\times C}$ are trainable weights. We apply a padding of \( \lfloor \frac{p}{2} \rfloor \) and a stride of 1 to preserve the same temporal dimension \( T \) across layers. The goal of this operation is to process the temporal speed information and translate it into volume-related information on a consistent temporal scale. Then, a DGAT layer is applied along the spatial dimension following \eqref{eq:dgat0}:
\begin{align}
h_{v,t}^{(k-1, 2)} = \sigma \left( 
    \sum_{u \in \mathcal{N}_1(v)}
    \alpha_{vu}^{\text{dir}\,(k)} 
     \left( W^{\text{dir} \, (k)} \, h_{u,t}^{(k-1,1)} \right)
\right),
\end{align}
where \( h_{v,t}^{(k-1, 2)} \in \mathbb{R}^C \). Finally, a second symmetric 1D-CNN is applied again along the temporal dimension to complete the factorized spatio-temporal block:
\begin{align} 
h_{v,t}^{(k, 0)} = \sigma \left( \sum_{i=0}^{p-1} W_i^{(k, 2)} \, h^{(k-1, 2)}_{v, t + i - \left\lfloor \frac{p}{2}\right\rfloor}\right) .\label{eq:stgnn3}
\end{align}
After applying $K$ factorized spatio-temporal blocks, we apply a fully connected layer to reduce the temporal dimension to 1 and obtain a final spatio-temporal embedding $ hp \in \mathbb{R}^{|\mathcal{N}_K(v_l)| \times C}$.

\subsection{Processing Static Descriptors with a Spatial Branch}
The spatial branch processes the static descriptor tensor \( F_{\mathcal{N}_K(v_l)} \in \mathbb{R}^{|\mathcal{N}_K(v_l)| \times C_f} \), which contains \( C_f \) static features for each road segment in the input subgraph. To capture time-independent spatial dependencies based on geospatial attributes, we apply $K$ DGAT layers over these features, for all nodes \( v \in \mathcal{N}_K(v_l) \), using update rule \eqref{eq:dgat0}:
\begin{align}
h_{v}^{(k)} = \sigma \left( 
    \sum_{u \in \mathcal{N}_1(v)} 
    \alpha_{vu}^{\text{dir}\,(k)} 
    \left( W^{\text{dir}\,(k)} \, h_{u}^{(k-1)} \right) 
\right),
\end{align}
where \( h_v^{(k)} \in \mathbb{R}^{C} \) is the hidden representation of node \( v \) at layer  \( k \in \{1, \ldots, K\} \). Applying these $K$ layers yields the final spatial embedding \( hf \in \mathbb{R}^{|\mathcal{N}_K(v_l)| \times C} \).

\subsection{Concatenation and Final Volume Profile Estimation}

In the final stage of the architecture, we concatenate the complementary embeddings $hp$ and $hf$ produced by both branches along the hidden channel dimension, resulting in a global representation \( H \in \mathbb{R}^{|\mathcal{N}_K(v_l)| \times 2C} \). This combined embedding is then passed through two fully connected layers with output dimensions of $C$ and $T'$, respectively, to generate the predicted volume profiles for each node in the subgraph \( G_{\mathcal{N}_K(v_l)} \), stored in the tensor \( \hat{Q}_{\mathcal{N}_K(v_l), d} \in \mathbb{R}^{|\mathcal{N}_K(v_l)| \times T'} \).

\section{Experiments}
\subsection{Dataset Construction}
Most publicly available traffic datasets do not support analysis on a topological road network graph. Moreover, very few datasets provide both speed and volume data simultaneously (e.g., PEMS-D4, PEMS-D8 \cite{b24}), and even fewer offer access to detailed topological descriptors. To address these limitations, we construct a custom dataset based on traffic and geospatial data collected in Lyon, France, gathering all necessary data to evaluate our model.

\subsubsection{Volume Sensor Data}
We collect traffic volume measurements from 653 sensors installed across various road types in the Lyon metropolitan area, covering the entire year of 2021. The volume values are aggregated at an hourly resolution, with an overall missing data rate of approximately 9\%.

\subsubsection{HERE Maps Static Descriptors and Speed Patterns}
To construct our input tensors $F$ and $P$, we use geospatial data from the HERE Maps platform. This includes both static road descriptors and speed profiles for all road segments in the network. We retain seven static features tied to traffic flow state, such as the speed limit or the number of lanes, as shown in Table \ref{Tab1}. The Functional Class is an artificial HERE Maps attribute that categorizes roads based on their role and importance within the transportation network, ranging from major highways and intercity roads to local streets and secondary routes. In addition, HERE maps provides average speed profiles for each day of the week, resulting in seven distinct temporal profiles for any road segment in the network. These speed profiles are 96-element sequences, where each element represents the average speed over a 15-minute interval. Examples of speed profiles for two labeled nodes are given in Figure \ref{fig:patterns}, along the corresponding measured volume profiles.

\begin{table}[h]
    \centering
    \caption{Selected static descriptors and corresponding operations used for the primal graph links aggregation}
    \label{Tab1}
    \renewcommand{\arraystretch}{1.1}
    \begin{tabularx}{\columnwidth}{@{}l|
        >{\centering\arraybackslash}p{1.7cm}|
        >{\centering\arraybackslash}p{1.9cm}|
        >{\centering\arraybackslash}p{1.7cm}@{}}
        \hline
        \textbf{Features} & \textbf{Variable Type} & \textbf{Range (unit)} & \textbf{Aggregation} \\
        \hline
        Speed Limit & Discrete & 10 -- 130 (km/h) & MIN \\ 
        Number of Lanes & Discrete & 1 -- 4 (-) & MIN \\
        Link Length & Continuous & 2 -- 2127 (m) & SUM  \\
        Free Flow Speed & Continuous & 10 -- 120 (km/h) & MEAN \\
        Curvature & Continuous & 0 -- 1000 (1/m) & MEAN \\
        Slope Percent & Continuous & -100 -- 100 (\%) & MEAN \\
        Functional Class & Categorical & 1 -- 5 (-) & MIN \\
        \hline
    \end{tabularx}
\end{table}

\begin{figure}[h]
    \centering
    \setlength{\tabcolsep}{1pt} 
    \renewcommand{\arraystretch}{0.5} 
    \begin{tabular}{cc}
        \includegraphics[width=0.5\columnwidth]{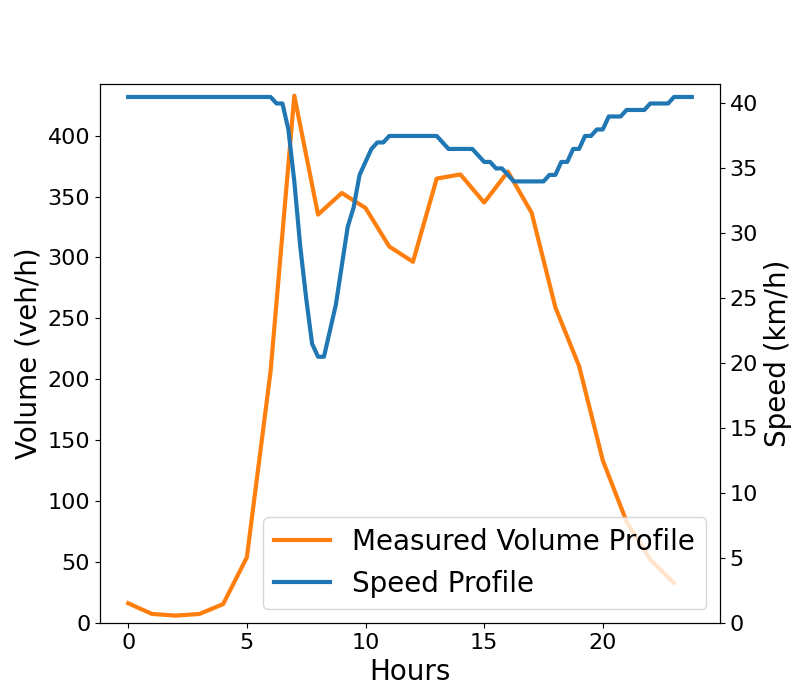} &
        \includegraphics[width=0.5\columnwidth]{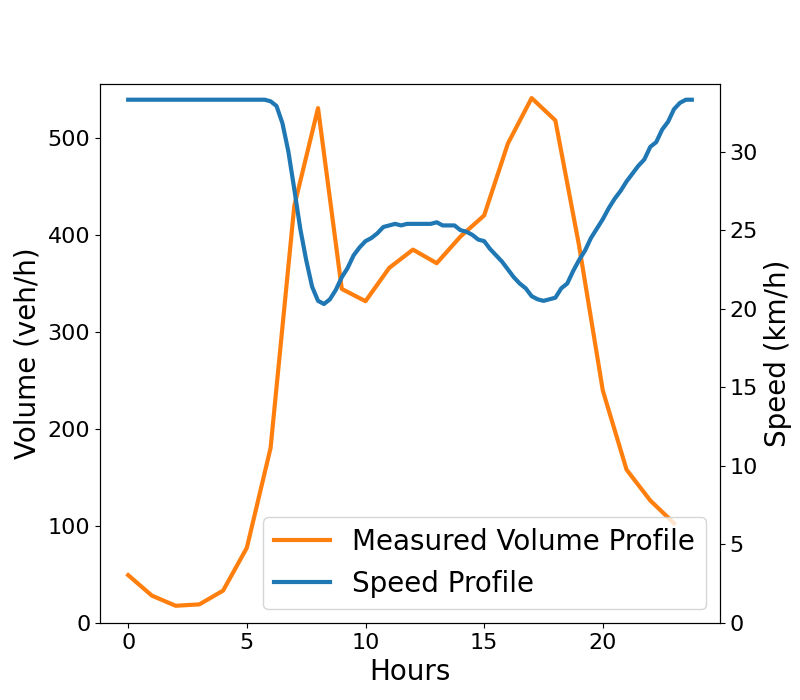} \\
    \end{tabular}
    \caption{Examples of speed profiles associated with the measured volume profiles for two specific days and labeled road sections.}
    \label{fig:patterns}
\end{figure}

\subsubsection{Oriented Topological Dual Graph Construction}
We reconstruct Lyon’s topological road network using HERE Maps geospatial data, providing link-level information, which are road segments split based on attribute changes rather than physical characteristics. These links vary in length and often don’t represent meaningful road sections, leading to inconsistencies in neighborhood definitions across the graph. For example, two road sections with a 3-hop neighborhood can differ significantly in terms of the number of intersections they cover, distorting the spatial meaning of the neighborhood.

This irregular segmentation creates two challenges for our approach. First, inconsistent spatial scales across neighborhoods affect dependency modeling. Second, deep Graph Neural Networks are prone to vanishing gradients and over-smoothing which can degrade performance, while reaching sufficiently distant neighborhoods remains essential, as traffic volume is not expected to vary significantly within links between two intersections. To address these challenges, we aggregate each set of links connecting two intersections into a single unified road segment, assuming traffic volume is stable at an hourly scale and neglecting exogenous flows. As a result, we aggregate the corresponding static features, when necessary, using appropriate aggregation operations, shown in Table \ref{Tab1}, to create new features for the unified segment. Speed profiles are also averaged over all aggregated links in a new segment.

\begin{figure}
    \centering
    \includegraphics[width=1\linewidth]{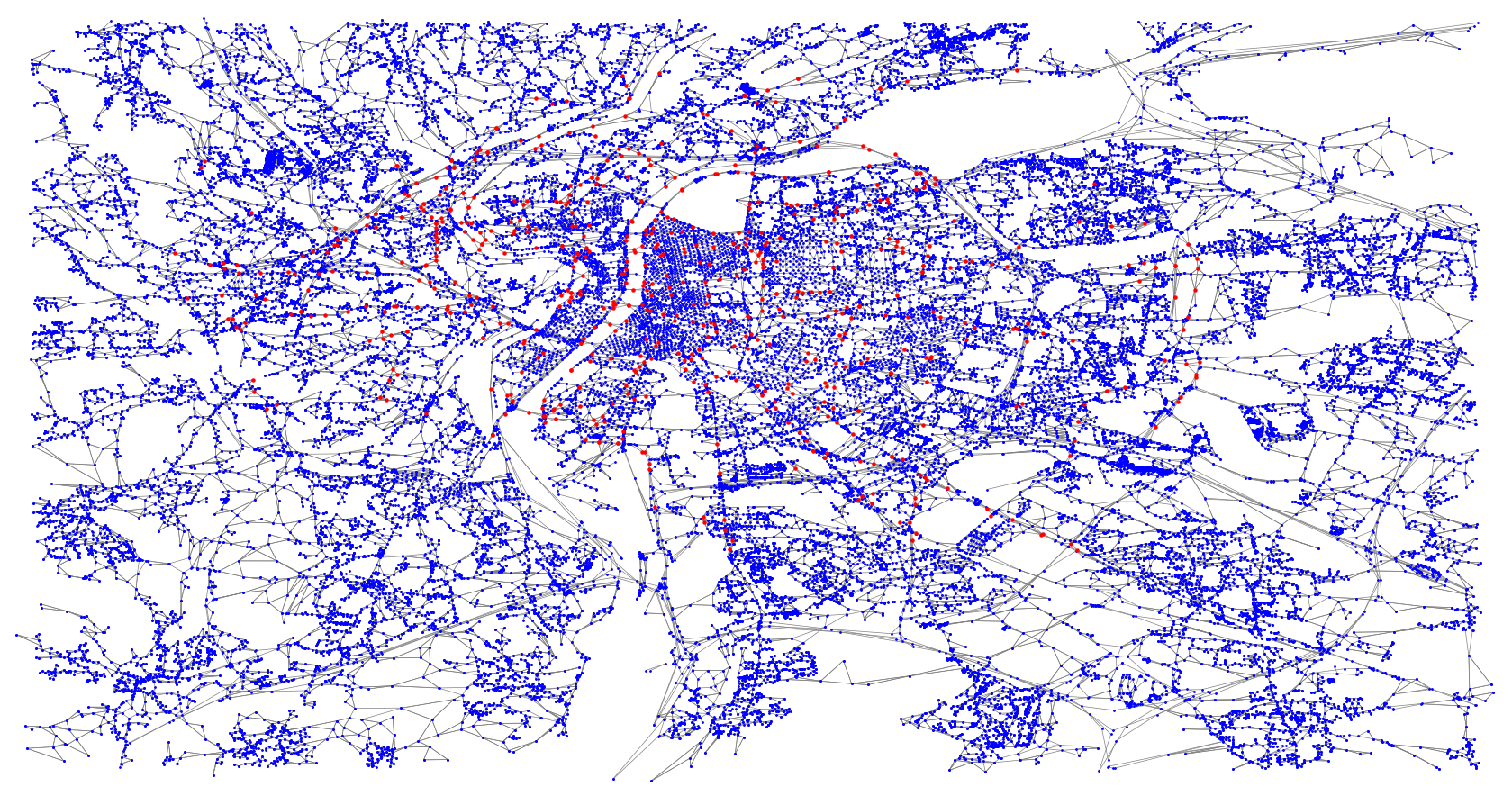}
    \caption{Created dual graph network of Lyon with sensors as red dots.}
    \label{Fig:dual}
\end{figure}

After aggregating the primal graph, we use HERE Maps data to extract traffic rules and generate the corresponding dual graph, shown in Figure~\ref{Fig:dual}. The graph assigns a distinct node for each direction on a road section and is subsequently oriented based on the traffic flows in the primal graph to model all possible maneuvers, making it compatible with our DGAT layers.

\subsubsection{Dataset Creation}

Following the sample definition introduced earlier \eqref{Eq:sample}, we construct our Lyon Dataset using all available data, considering 2-hop neighborhoods as a balanced trade-off between model complexity, computation time, and the spatial scale necessary to capture local traffic behavior. Although our approach is applicable to day-specific input probe speed data for estimating daily volume profiles, available speed data is averaged for each weekday throughout the whole year of 2021. As we lack additional temporal features with sufficient descriptiveness to support more fine-grained temporal predictions, we average, for each of the 653 sensors, all available volume profiles from the 365 days of 2021, grouped by weekday. We then assess our model's ability to estimate typical weekday volume profiles for each road segment in the network. However, we also evaluate the model on raw volume data to better understand the impact of daily temporal variations. The dataset then contains 4159 unique samples with averaged volumes and 216 885 unique samples with raw volumes.

\subsubsection{Training Nodes Representativeness}

\begin{figure}
    \centering
    \includegraphics[width=1\linewidth]{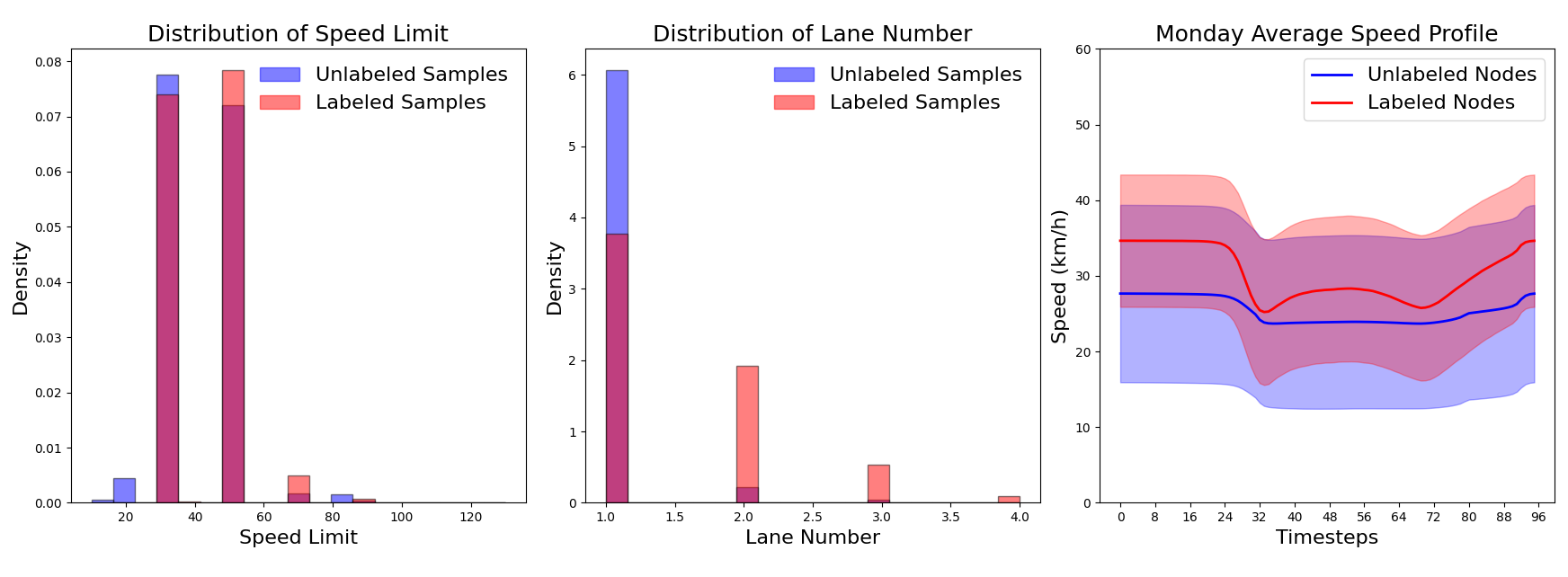}
    \caption{Input features representativeness of training nodes. From left to right: speed limit distributions, lane number distributions, and average Monday speed profiles (with standard deviation per time step), for both unlabeled and labeled nodes in the complete dual graph.}
    \label{Fig:repres}
\end{figure}

By creating this dataset with such sampling, our approach aims to learn a local transformation on the dual road graph using the associated features to estimate the target volume profile. Since we adopt a supervised approach, the distribution of training node features is crucial for the spatial generalization of our model to previously unseen local neighborhoods. This places our work within the domain of out-of-distribution generalization. Our method is designed to be highly generalizable, and the datasets used could be extended to other sensors (and other cities) to enhance the variability and consistency of this feature distribution, which may be limited in our case. To assess the extrapolation feasibility within the same network, we present in Figure \ref{Fig:repres} the normalized distributions of some key input features for both unlabeled nodes $V_n$ and labeled nodes $V_l$. While the labeled data may not be perfectly representative, the differences between distributions appear small enough to expect reliable inference on the majority of nodes in the complete graph of Lyon.

\subsection{Experimental Setting}

After each model forward pass, we extract the estimated volume profile \( \hat{Q}_{v_l, d} \in \mathbb{R}^{T'} \) of the target node \( v_l \) for day \( d \) from the resulting tensor \( \hat{Q}_{\mathcal{N}_K(v_l), d} \). We then compute the Huber loss (with threshold \( \delta = 50 \)) between this estimate and the corresponding ground-truth profile \( Q_{v_l, d} \in \mathbb{R}^{T'} \), both rescaled to their original units. Since both profiles contain \( T' \) time steps, the loss is defined as the mean of the element-wise Huber losses across the temporal dimension:
\begin{align*}
L_\delta(Q_{v_l, d}, \hat{Q}_{v_l, d}) =
\frac{1}{T'} \sum_{t=1}^{T'} 
\begin{cases}
\frac{1}{2} a_t^2 & \text{if } |a_t| \leq \delta \\
\delta \left(|a_t| - \tfrac{1}{2}\delta\right) & \text{otherwise}
\end{cases},
\end{align*}
where \( a_t = Q_{v_l,t, d} - \hat{Q}_{v_l,t, d} \) is the error at time step \( t \).

Experiments are conducted on a workstation equipped with an \textbf{NVIDIA GeForce RTX 4070 GPU} and an \textbf{Intel i5-5600KF CPU}. Model hyperparameters are manually tuned. The input speed profiles have a temporal length of $T = 96$ (15-minute resolution), while volume profiles are limited to $T' = 24$ (hourly resolution). Both the graph neighborhood depth and the depths of the spatio-temporal and spatial branches are set to $K = 2$. We use $C_f = 7$ static descriptors and the number of considered work days is $D = 7$, as we average volume profiles for each weekday to align with the daily resolution of input speed profiles. All hidden dimensions are fixed at $C=64$ across the model. We use $LeakyReLU()$ activation functions with negative slopes of $0.2$ for our DGAT layers and $ReLU()$ for any other layers. The 1D-CNN layers use a kernel size of 9, while DGAT layers employ 4 attention heads. Dropout is applied after each layer, with dropout rates adjusted depending on the type of layer, ranging from 0.1 to 0.6. Batch size is set to 64.

To assess the spatial generalization ability of our model, we randomly select 20$\%$ of the sensor nodes as a validation set, ensuring they are completely excluded from the training process. Training is then operated on 522 labeled nodes using all available days, which is 7 in our experiments. Results are averaged over five different random seeds to reduce bias introduced by the validation nodes sampling. 

We use the following evaluation metrics: \textbf{RMSE}, \textbf{MAPE}, \textbf{GEH} \cite{b25}, and \textbf{\%GEH}$>$\textbf{5}. The GEH (Geoffrey E. Havers) statistic is particularly well-suited for comparing estimated and observed traffic volumes, and is widely used in transportation modeling. A common threshold considers predictions satisfactory when at most 15$\%$ of estimated values yield a GEH higher than 5.

Finally, to validate our architectural design, we conduct a series of ablation studies, testing the impact of: 
\begin{itemize}
    \item removing the speed spatio-temporal branch (1),
    \item removing the static descriptor spatial branch (2),
    \item removing the neighborhood information (3),
    \item removing the multi-branch fusion mechanism (4),
    \item using standard GAT layers on an undirected graph (5),
    \item evaluating on raw volume profiles (6).
\end{itemize}

\subsection{Experimental Results}

\begin{figure}
    \centering
    \includegraphics[width=1\linewidth]{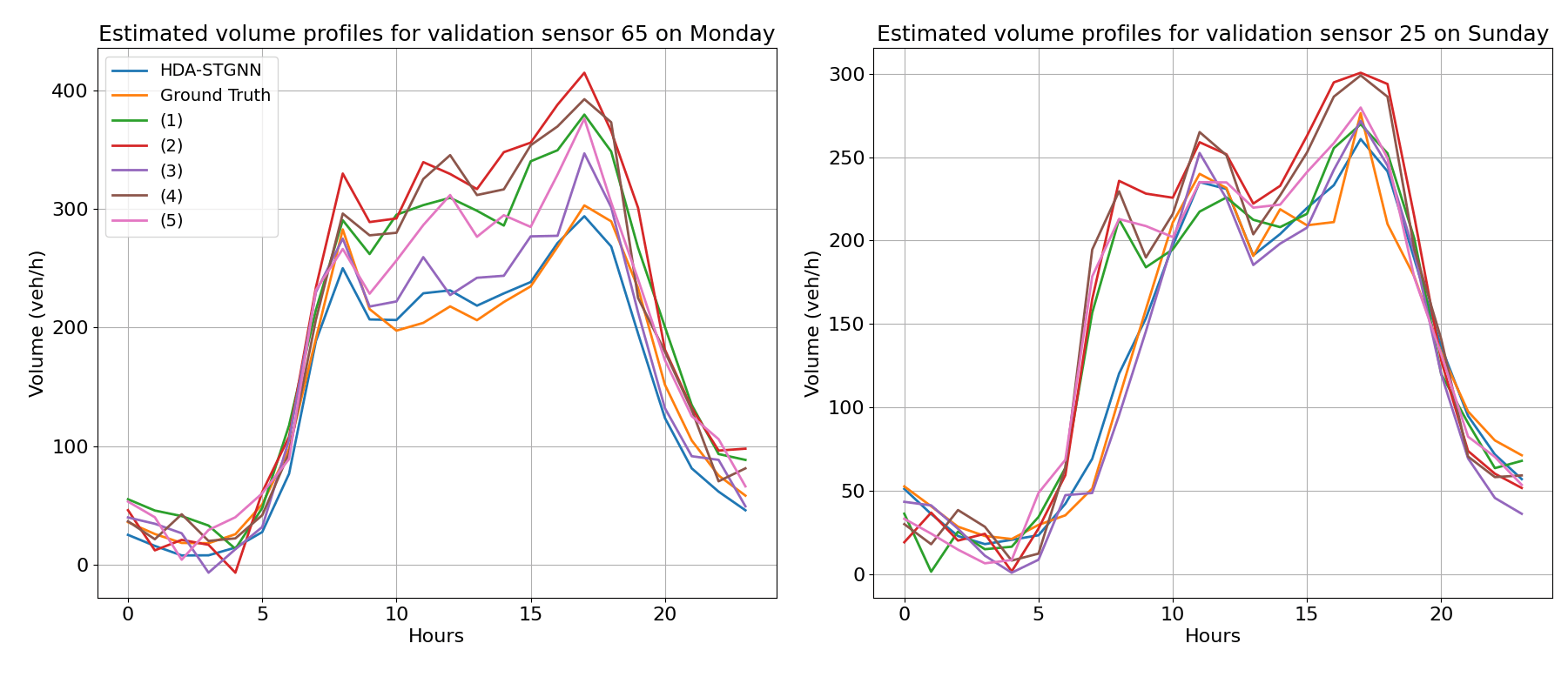}
    \caption{Examples of estimated traffic volume profiles by HDA-STGNN and its ablation variants on two specific roads and days.}
    \label{fig:flows}
\end{figure}

Table \ref{Tab2} presents the performance of HDA-STGNN and all ablation studies on the constructed Lyon Dataset, while Figure \ref{fig:flows} shows examples of estimated flow profiles. Our model achieves the best results with a mean GEH of 6.09, which constitutes a promising outcome for such a challenging estimation task.
The ablation studies provide a detailed understanding of the role played by each component of our architecture. Ablations (1) and (2) clearly show the importance of jointly using static road descriptors and speed profiles as input features. In particular, omitting static attributes leads to a significant degradation in performance, with a mean GEH rising to 8.18. This confirms that static features offer essential structural information about road capacity and nominal volume profiles, whereas speed profiles provide more granular information on road usage and daily flow fluctuations. Without the structural context, the model struggles to generalize to unseen roads, as speed signals alone lack the semantic richness required for an effective spatial extrapolation. Ablation (4) further supports this point, showing that merging static and temporal inputs into a single spatio-temporal branch, by duplicating static features across the time dimension, yields suboptimal performance. This suggests that the hybrid two-branch architecture better captures, in a complementary manner, the fundamentally different nature of static and dynamic inputs.

\begin{table}[h]
    \centering
    \caption{Performance comparison of HDA-STGNN and ablation studies on the created Lyon Dataset.}
    \label{Tab2}
    \renewcommand{\arraystretch}{1.1}
    \begin{tabularx}{\columnwidth}{@{}l|>{\centering\arraybackslash}p{1.2cm}|>{\centering\arraybackslash}p{1.2cm}|>{\centering\arraybackslash}p{1.2cm}|>{\centering\arraybackslash}p{1.7cm}@{}} 
        \hline
        \textbf{Model} & \textbf{RMSE} & \textbf{MAPE} & \textbf{GEH} & \textbf{\%GEH} $>$ \textbf{5} \\
        \hline
        HDA-STGNN & \textbf{177.98} & \textbf{52.36 }& \textbf{6.09} & \textbf{46.61 }\\
        Ablation (1) & 197.61 & 52.86 & 6.53 & 49.22 \\ 
        Ablation (2) & 241.56 & 72.11 & 8.18 & 57.94 \\
        Ablation (3) & 205.98 & 58.43 & 6.78 & 52.01 \\
        Ablation (4) & 228.15 & 67.74 & 7.93 & 56.32 \\
        Ablation (5) & 190.44 & 53.21 & 6.46 & 50.86 \\
        Ablation (6) & 210.76 & 60.88 & 6.82 & 52.47 \\
        \hline
    \end{tabularx}
\end{table}

Ablation (3) highlights the value of exploiting local topological structure through Graph Neural Networks. When the Directed Graph Attention (DGAT) layers are replaced with simple linear transformations, the model loses its capacity to incorporate contextual information from neighboring roads, which proves valuable for refining volume predictions. Similarly, ablation (5) demonstrates the benefit of using directed attention mechanisms over standard graph attention, suggesting that the model learns to interpret directional dependencies in traffic dynamics more effectively when spatial flow orientation is explicitly modeled.

Ablation (6) explores the impact of fine-grained temporal variations in traffic volume. Although our model focuses on estimating daily average volume profiles, this experiment reveals the influence of intra-day variability and suggests that incorporating richer temporal descriptors (e.g. weather, calendar events, traffic disruption) could further improve accuracy in future work. Nevertheless, our current approach achieves robust estimates that capture meaningful traffic patterns across the network.

Model performance is also subject to variation across samples. In particular, we observe that predictions degrade for road segments whose feature profiles diverge significantly from the training distribution, revealing limitations in out-of-distribution generalization. This issue could potentially be mitigated by expanding the training dataset to include a more diverse set of local topologies and associated features. Overall, the model effectively captures volume magnitudes on most roads and refines temporal patterns using speed inputs, with more or less difficulties. For some segments, the model significantly underestimates or overestimates peak intensities, particularly during morning rush hours (e.g., 8–9a.m.). Many of these segments are almost always in a fluid regime, and the model may lack the necessary features to infer true demand levels, information not reflected by constant speed profiles and insufficiently compensated by neighborhood context.

\begin{figure}
    \centering
    \includegraphics[trim=0cm 0.8cm 0cm 0cm, clip, width=1\linewidth]{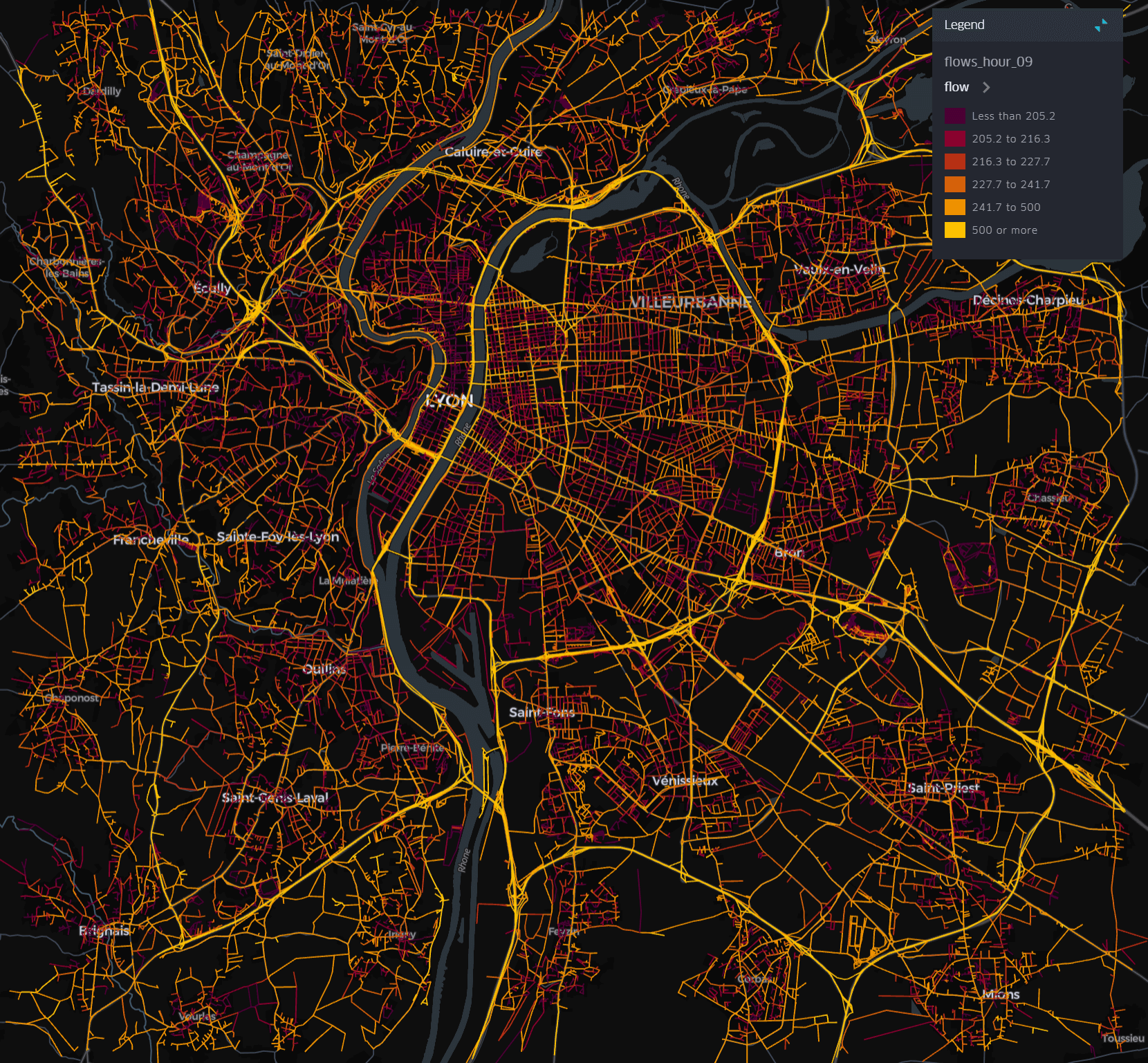}
    \caption{Network-wide hourly volume estimation for Tuesdays at 9a.m.. Roads are categorized by volume intensity, with yellow representing the highest volumes and purple representing the lowest ones.}
    \label{fig:fullpred}
\end{figure}

Knowing that the final objective of this study is to estimate traffic volume on a complete road network using a transformation learned from a small sample of labeled roads, we perform inference for a specific weekday, applying this transformation to all nodes in the Lyon road graph and visualize the results for 9a.m. in Figure \ref{fig:fullpred}. Although a quantitative evaluation of unlabeled roads is not possible, the results demonstrate good overall consistency in the network volume distribution, with higher values on major arteries and lower values on smaller roads. However, some limitations are apparent, particularly a lack of flow conservation at certain intersections, which could be mitigated by incorporating regularization techniques to enhance the local coherence of predictions. Nevertheless, the results are promising and provide an effective way to estimate the global traffic state on the complete network for any given input speed data corresponding to a specific day.

\section{Conclusion and Future Work}

In this paper, we proposed a novel deep learning framework, HDA-STGNN, for network-wide traffic flow estimation. Our model combines directed graph attention networks with temporal convolution in a hybrid architecture, specifically designed to leverage both speed profiles and static road descriptors within a dual graph representation of the road network. Experimental results demonstrate that our approach achieves promising performance, using a real-world dataset constructed from traffic data collected in the city of Lyon. These results highlight the model’s strong potential in addressing the challenge of spatial generalization to unknown roads. Future work will focus on increasing the variability and diversity of the training dataset, as well as incorporating physical constraints to further enhance the realism and consistency of the predictions. Additionally, thanks to its inductive design, our approach could be transferred to unseen road networks and applied to any day with available input data, including sensor-free areas, enabling broader applications in traffic analysis and smart city planning.

\section{Acknowledgement}
This work is funded by the French National Research Agency as part of the Mob Sci-Dat Factory project (ANR-23-PEMO-0004) under the France 2030 program.

\end{document}